\documentclass[10pt,twocolumn,letterpaper]{article}

\usepackage[pagenumbers]{iccv} % To force page numbers, e.g. for an arXiv version

\usepackage{multirow}  
\usepackage{makecell}
\usepackage{tablefootnote}

\definecolor{iccvblue}{rgb}{0.21,0.49,0.74}
\usepackage[pagebackref,breaklinks,colorlinks,allcolors=iccvblue]{hyperref}

\def\paperID{12394} % *** Enter the Paper ID here
\def\confName{ICCV}
\def\confYear{2025}

\title{ProGait: A Multi-Purpose Video Dataset and Benchmark for Transfemoral Prosthesis Users}

\author{Xiangyu Yin\hspace{0.3in} Boyuan Yang\hspace{0.3in} Weichen Liu\hspace{0.3in} Qiyao Xue \\Abrar Alamri\hspace{0.3in} Goeran Fiedler\hspace{0.3in} Wei Gao\\
University of Pittsburgh\\
{\tt\small \{eric.yin, by.yang, weichenliu, qiyao\_xue, aba114, gfiedler, weigao\}@pitt.edu}
}

\begin{document}
\maketitle

\begin{abstract}
	\vskip -0.15in
	Prosthetic legs play a pivotal role in clinical rehabilitation, allowing individuals with lower-limb amputations the ability to regain mobility and improve their quality of life. Gait analysis is fundamental for optimizing prosthesis design and alignment, directly impacting the mobility and life quality of individuals with lower-limb amputations. Vision-based machine learning (ML) methods offer a scalable and non-invasive solution to gait analysis, but face challenges in correctly detecting and analyzing prosthesis, due to their unique appearances and new movement patterns. In this paper, we aim to bridge this gap by introducing a multi-purpose dataset, namely ProGait, to support multiple vision tasks including Video Object Segmentation, 2D Human Pose Estimation, and Gait Analysis (GA). ProGait provides 412 video clips from four above-knee amputees when testing multiple newly-fitted prosthetic legs through walking trials, and depicts the presence, contours, poses, and gait patterns of human subjects with transfemoral prosthetic legs. Alongside the dataset itself, we also present benchmark tasks and fine-tuned baseline models to illustrate the practical application and performance of the ProGait dataset. We compared our baseline models against pre-trained vision models, demonstrating improved generalizability when applying the ProGait dataset for prosthesis-specific tasks. Our code is available at \url{https://github.com/pittisl/ProGait} and dataset at \url{https://huggingface.co/datasets/ericyxy98/ProGait}.
\end{abstract}
 
\vspace{-0.2in} 
\section{Introduction}
\vspace{-0.05in}
Prosthetic legs have been a transformative solution in clinical rehabilitation, allowing individuals with lower-limb amputations the ability to regain mobility and improve their quality of life. The success of a prosthetic leg relies heavily on its alignment on individual users that is crucial to comfort, stability, and efficient movement \cite{zahedi1986alignment}, and gait analysis plays a pivotal role in achieving optimality in such alignment, by providing detailed insights into the user's individualized walking patterns \cite{schmalz2002energy,zhang2019effect,cardenas2022effect} and further indicating fitness of the prosthetic leg on each user. Moreover, gait analysis also helps clinicians identify misalignments, assess biomechanical efficiency, and make precise adjustments to the prosthesis, ultimately enhancing mobility and functionality for the user.

\begin{figure}[t]
%	\vskip 0.1in
	\begin{center}
		\includegraphics[width=\columnwidth]{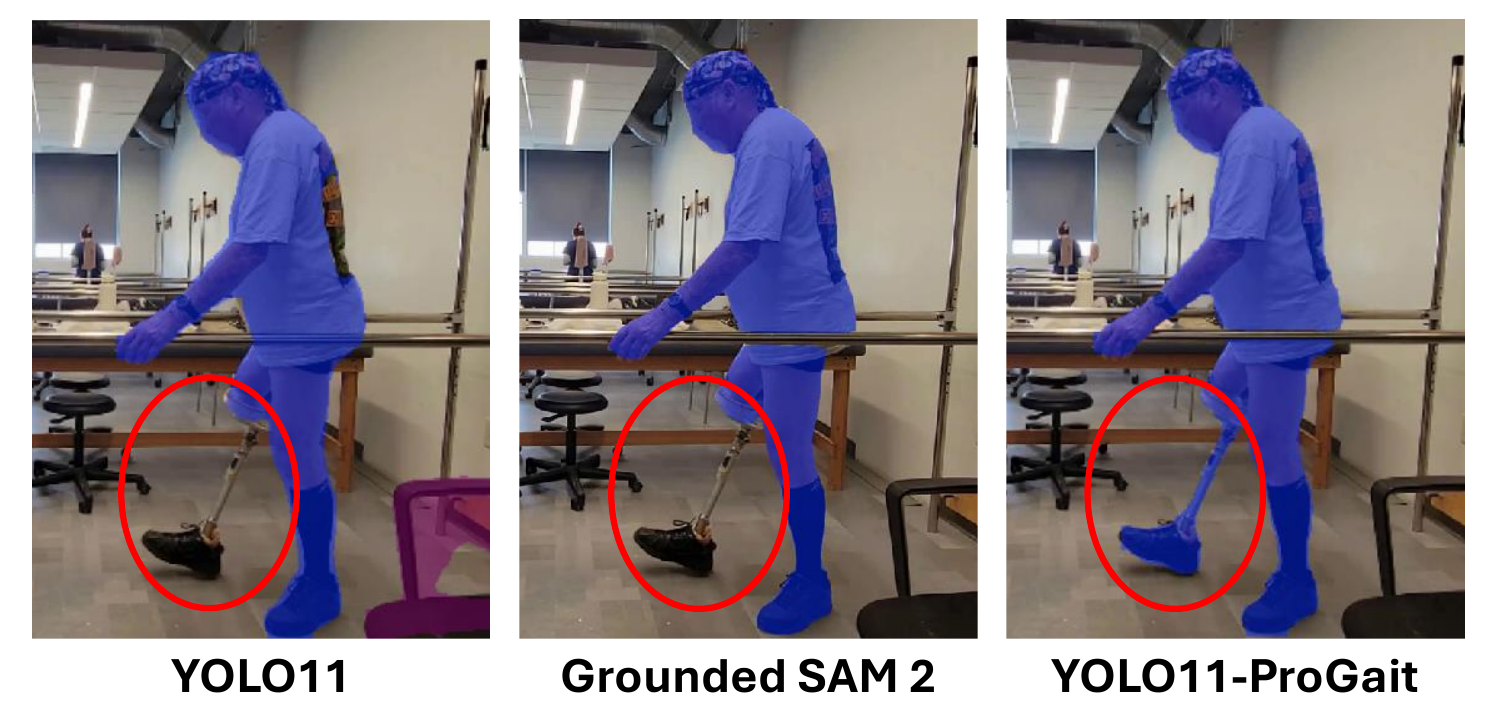}
		\vskip -0.1in
		\caption{Limitation of current vision-based ML models on detecting the whole body of users with prosthetic legs. Left: Pre-trained YOLO11 model \cite{yolo11_ultralytics}; Middle: Grounded SAM2 model \cite{ren2024grounded} with text prompt ``a person with prosthetic leg''; Right: YOLO11 model fine-tuned on our ProGait dataset}
		\label{fig:sota_limitation}
	\end{center}
	\vskip -0.35in
\end{figure}

\begin{figure*}[ht]
	\vskip -0.2in
	\begin{center}
		\includegraphics[width=0.95\textwidth]{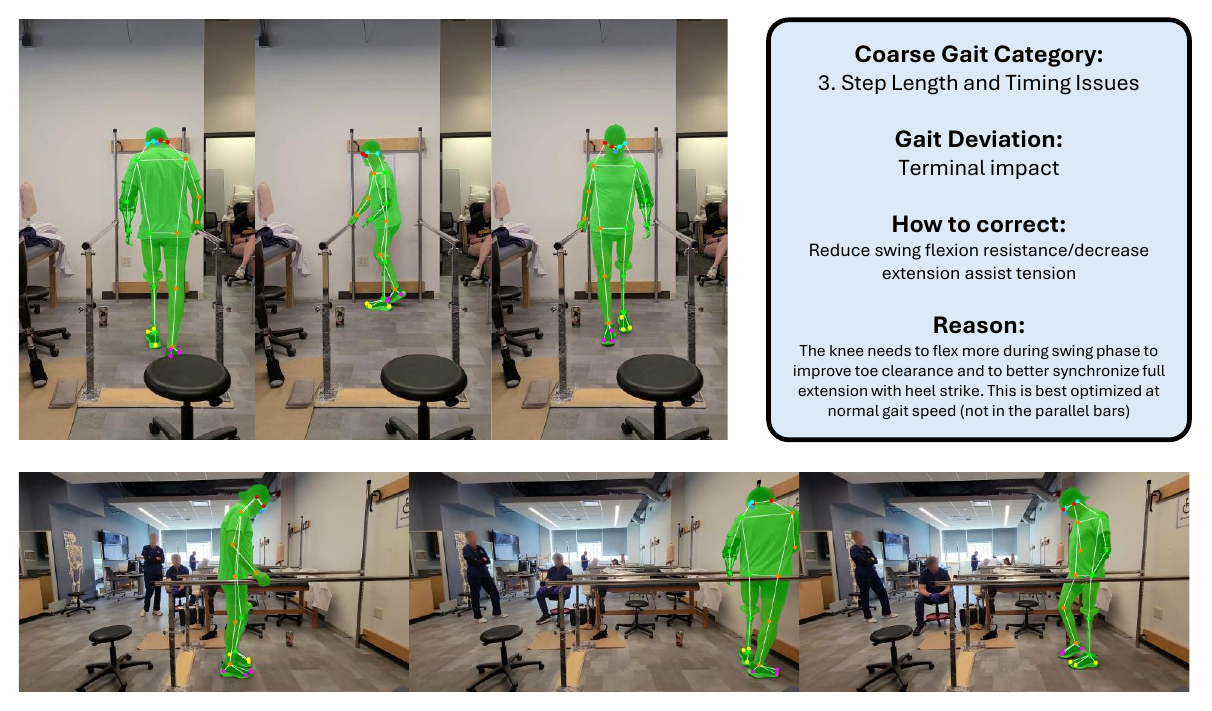}
		\vskip -0.1in
		\caption{Examples of video frames and annotations from the ProGait dataset. \textbf{Top-left}: Frontal view with segmentation masks and pose keypoints. \textbf{Bottom}: Sagittal view with segmentation masks and pose keypoints. \textbf{Top-right}: Textual descriptions for gait analysis. This sample was collected in the inside parallel bar scenario. Additional samples are available in the supplementary material.}
		\label{fig:examples}
	\end{center}
	\vskip -0.2in
\end{figure*}

Traditionally, gait analysis relied on specialized motion capture systems or embedded on-body motion sensors to measure biomechanical parameters \cite{ceseracciu2014comparison,pfister2014comparative,cloete2008benchmarking,redmon2016you}. While these approaches provide high-fidelity data, they are often expensive, intrusive, and limited to controlled environments, thereby constraining their accessibility and practicality in real-world applications at scale. 

Vision-based machine learning (ML) approaches have recently emerged as a promising alternative \cite{singh2018vision, khan2013computer}, and can extract spatio-temporal information about human motion directly from video data, enabling non-invasive, scalable, and cost-effective gait analysis. However, current vision models face significant limitations in accurately detecting and analyzing individuals with transfemoral prosthetic legs \cite{cimorelli2024validation}, as shown in Figure \ref{fig:sota_limitation}. This is because current models are mostly trained on large datasets that only contain able-bodied individuals and hence fail to account for the unique appearance and movement patterns of prosthetic limbs. This gap in model performance not only hampers accurate gait analysis, but also impedes many downstream tasks such as rehabilitation assessment \cite{esquenazi2014gait} and prosthesis optimization \cite{price2019design} that are important to human well-beings.

To address this challenge, in this paper, we present a new multi-purpose video dataset, namely \emph{ProGait}, that depict the presence, contours, poses, and gait patterns of human subjects with transfemoral prosthetic legs. As shown in Figure \ref{fig:examples}, our dataset is designed to mainly support three important tasks in vision-based analysis: \emph{1)} Video Object Segmentation (VOS), \emph{2)} 2D Human Pose Estimation (HPE), and \emph{3)} Gait Analysis (GA). By providing high-quality annotations and diverse scenarios, this dataset enables evaluation and fine-tuning of vision models, to improve their performance on prosthesis-specific tasks, as well as in-depth studies of prosthesis dynamic alignment. More video samples are illustrated in Appendix \ref{sec:video_samples}.

\begin{figure}[t]
%	\vskip -0.1in
	\begin{center}
		\includegraphics[width=1\columnwidth]{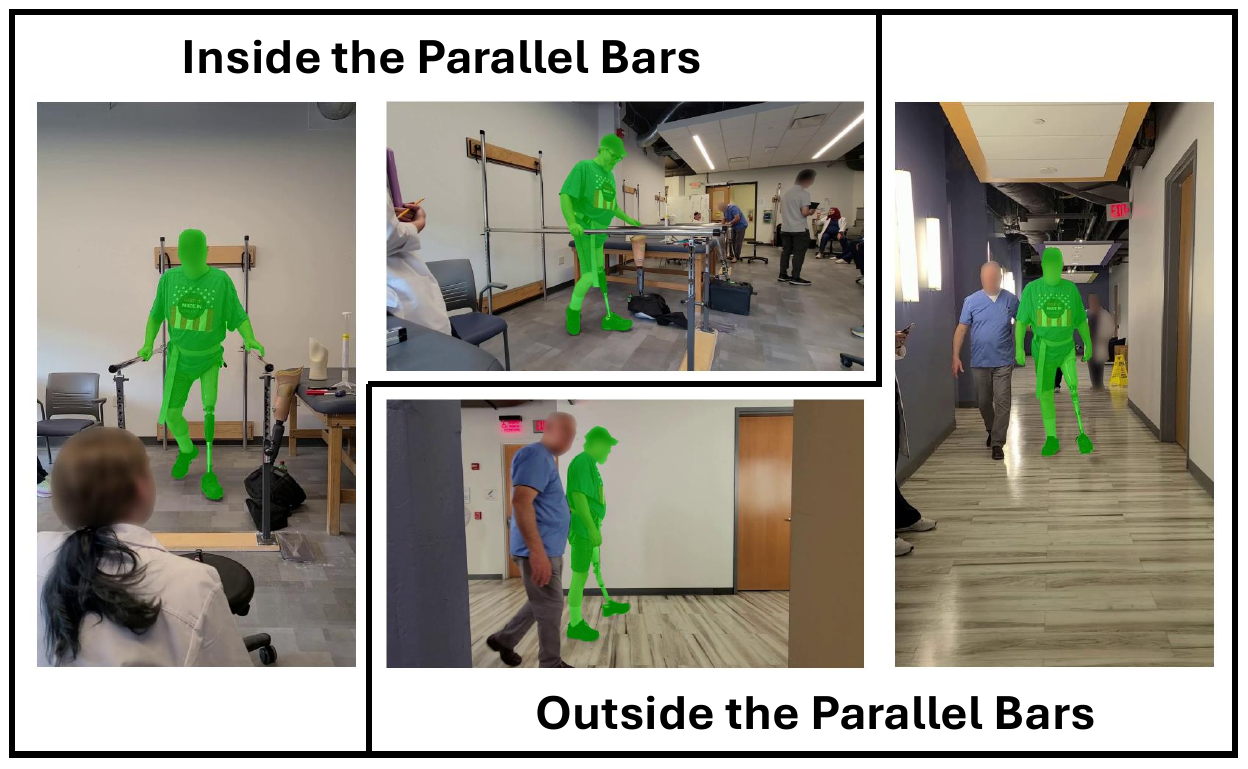}
		\vskip -0.1in
		\caption{The two scenarios}
		\label{fig:two_scenarios}
	\end{center}
	\vskip -0.2in
\end{figure}

We collected 412 video clips from four above-knee amputees, when testing multiple newly-fitted prosthetic legs through walking trials. The videos encompasses the following to primary scenarios as shown in Figure \ref{fig:two_scenarios}:

\begin{enumerate}
	\vspace{0.03in}
	\item \textbf{Walking \emph{inside the parallel bars} without external assistance}: Subjects navigate the parallel bars independently, focusing on balance and stability.
	\vspace{0.03in}
	\item \textbf{Walking in hallways \emph{outside the parallel bars} with assistance}: Subjects ambulate in open indoor spaces with support from others, simulating real-world walking conditions.\footnote{The assistants do not have direct contact or provide any physical assistance during the data collection. Their presence was solely for safety assurance, and they intervene only in the event of a potential fall.}
	\vspace{0.03in}
\end{enumerate}

Each walking trial includes both frontal and sagittal views, providing comprehensive perspectives for analysis. To ensure diversity and generalizability, the trials on each subject involve various types and configurations of prosthetic legs, different background contexts and lighting conditions, and heterogeneous presence of other human individuals. The dataset covers a diverse range of normal and abnormal gait patterns, each of which is accompanied by detailed textual descriptions from researchers in rehabilitation sciences and human engineering. These descriptions outline the correlations between abnormal gait deviations and the necessary corrective adjustments in order to regain normal gaits, as well as detailed reasons about why such adjustments are needed. All the video data collection, annotations and text descriptions have been approved by the institutional IRB.

We also provided benchmark tasks and fine-tuned YOLO11 \cite{yolo11_ultralytics} and RTMPose \cite{jiang2023rtmpose} models as baselines, to demonstrate how this dataset can be used to practically advance research in enhancing the accessibility and effectiveness of prosthetic solutions through vision models. We evaluate our baseline models against SOTA vision models in a zero-shot setup, with results showing that our baseline models outperform the SOTA models by 9\% in VOS task and 10-30\% in 2D HPE task. We also trained a top-down classification model for identifying 9 different gait patterns, and the overall accuracy is up to 81.2\% when taking sagittal views alone. These findings validate the effectiveness of our dataset in improving the generalizability of vision models for individuals with prosthetic legs.

%These findings prove that our dataset is sufficiently diverse and representative, validate the effectiveness of our dataset in improving the generalizability of vision models for individuals with prosthetic legs, and confirm that our dataset can serve as a robust foundation for training and evaluating vision-based models, ultimately advancing the field of assistive mobility technologies.

\section{Related Work}

\subsection{Gait Analysis and Vision Models}

Gait analysis has long been a cornerstone of understanding human locomotion, and used to rely on motion capture systems \cite{ceseracciu2014comparison,pfister2014comparative}, force plates \cite{budsberg1987force,liu2011mobile}, and embedded motion sensors \cite{cloete2008benchmarking,redmon2016you} to capture biomechanical parameters such as joint angles, ground reaction forces, and stride lengths. These methods, however, are often limited by their reliance on specialized equipment, controlled laboratory settings and high operational costs. 
%As a result, their accessibility for large-scale studies or real-world applications remains constrained.
%In recent years, vision-based gait analysis has emerged as a promising alternative to traditional methods. 
Vision-based models, such as YOLO \cite{redmon2016you} and the Segment Anything Model (SAM) \cite{kirillov2023segment}, allow more scalable, flexible and non-invasive gait assessment \cite{singh2018vision, khan2013computer}, by extracting the spatio-temporal features from video data and transforming gait analysis into typical vision tasks such as pose estimation, object tracking, and semantic segmentation.

Despite these advancements, current vision models often struggle to accurately detect and analyze individuals with prosthetic legs, primarily due to the unique appearance and movement patterns of prosthetic limbs \cite{cimorelli2024validation}. This limitation highlights the need for specialized datasets and tailored models to improve the generalizability of vision techniques for prosthetic users.

\subsection{The Existing Datasets}
Datasets have been available for vision-based gait analysis. Datasets like the Gait Abnormality in Video Dataset (GAVD) \cite{ranjan2024computer} and the Health\&Gait \cite{zafra2025health} dataset offer thousands of video sequences with detailed annotations, such as semantic segmentation and human pose. However, they lack representation of individuals with prosthetic limbs. %, limiting their applicability for prosthesis-specific gait analysis. 
A more specialized dataset focuses on kinematics and kinetics of 18 above-knee amputees walking at various speeds \cite{hood2020kinematic}. Although providing detailed motion capture data and offering valuable insights into the biomechanics of prosthetic users, it does not provide raw videos and is instead limited by the fact that participants only use their personal prosthetic devices.

In contrast, the ProGait dataset presented in this paper addresses these gaps by including individuals with prosthetic legs, when testing multiple prosthetic designs. This diversity allows for more comprehensive evaluations and aligns with the need for scalable vision-based approaches that do not rely on expensive motion capture systems.

\section{The ProGait Dataset}

The ProGait dataset primarily focuses on enabling the following three tasks: \emph{1)} Video Object Segmentation (VOS), \emph{2)} 2D Human Pose Estimation (HPE), and \emph{3)} Gait Analysis (GA). ProGait contains paired video clips that record the frontal and sagittal views of walking trials conducted by human subjects with prosthetic legs. Each video pair includes the corresponding annotations, such as bounding boxes, segmentation masks and pose keypoints of the subject, accompanied by descriptive gait assessments in text. These comprehensive annotations aim to support multi-task learning and fine-tuning of vision-based models. A sample of such a video pair and associated annotations and text descriptions can be found in Figure \ref{fig:examples}.

\subsection{Video Data Collection}

The video data was collected from four human subjects that are middle-aged/elderly above-knee amputees (with an average age of 61), representing a significant and specific sub-population within the amputee community - those who have lost their limbs due to vascular issues (e.g., diabetes-related complications leading to limb loss). This demographic constitutes a large portion of lower limb prosthesis users in the U.S. \cite{goodney2009national, goldberg2012effect, barnes2020epidemiology}. For this sub-population, safety and comfortable alignment are paramount, often prioritizing stability over high activity levels, especially with newly-fitted prostheses which are the focus of our data collection.

For the data collection, each subject tested multiple newly-fitted prosthetic legs and performed a series of walking trials both inside and outside the parallel bars as depicted in Figure \ref{fig:two_scenarios}. During the trials conducted outside the parallel bars in the open space, a healthcare professional accompanied the subjects to provide assistance and ensure their safety, particularly to prevent falls.

\begin{figure}[ht]
	\vskip -0.1in
	\begin{center}
		\includegraphics[width=0.9\columnwidth]{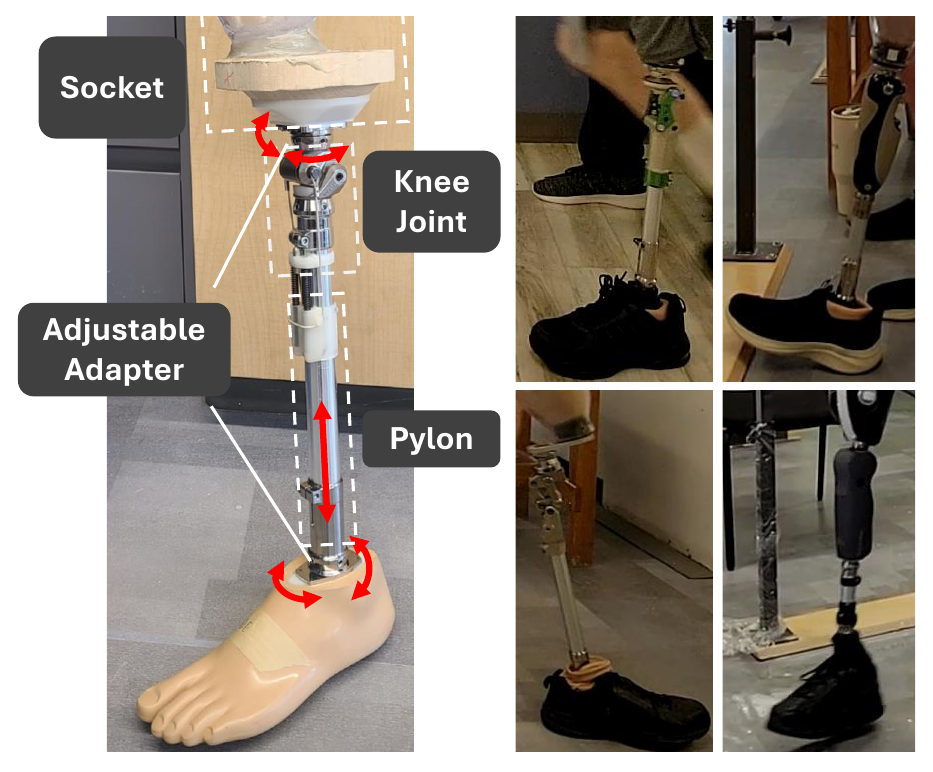}
		\vskip -0.1in
		\caption{Components of a transfemoral prosthetic leg (left) and different types of prosthetic legs (right)}
		\label{fig:prosthetic_legs}
	\end{center}
	\vskip -0.3in
\end{figure}

As shown in Figure \ref{fig:prosthetic_legs}, each prosthetic leg differs from each other, such that they have different types of knees (mechanical, hydraulic or computerized), different angles of knee and ankle joints, and different lengths of the pylon. Such differences affect the subject's gait patterns in different ways, and also result in very distinct visual appearances that make it hard for the vision models to recognize.

\begin{figure}[t]
%	\vskip -0.1in
	\begin{center}
		\includegraphics[width=0.95\columnwidth]{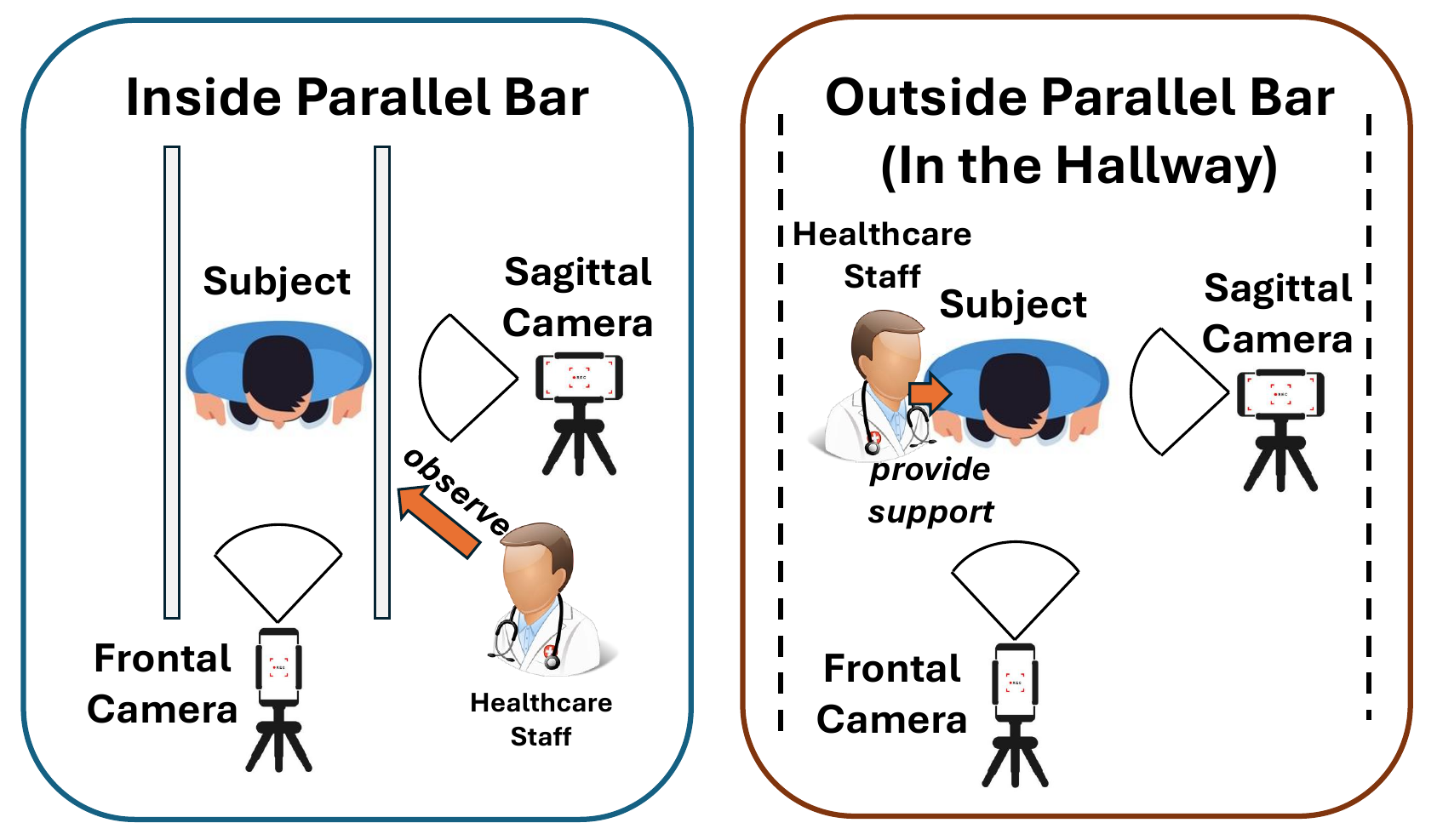}
		\vskip -0.1in
		\caption{Video Recording Setup}
		\label{fig:data_collection_setup}
	\end{center}
	\vskip -0.3in
\end{figure}

To capture the walking trials, we used two video cameras positioned at fixed locations to record the frontal and sagittal views of the subjects, as illustrated in Figure \ref{fig:data_collection_setup}. The videos were recorded at a resolution of 1920$\times$1080 and a frame rate of 30fps. Each walking trial consists of multiple round trips, with the best round trip selected as a single sample. In some cases, the healthcare staff may partially or fully obstruct the subject's view, particularly in the sagittal view during trials outside the parallel bars, as they walk alongside the subject to provide support. When this occurs, only a one-way trip is selected. The duration of selected walking trips lasts for 8-20 seconds in trials inside the parallel bars, and for 2-40 seconds for trials outside the parallel bars. During each trial, the subject walked approximately 5 meters inside the parallel bars and 10–15 meters outside the bars. The walking speed varied depending on the subject's comfort with the prosthesis.

In total, we collected 144 walking trials, resulting in 412 video clips contained in the dataset. Table \ref{tab:dataset_statistics} presents the basic distribution of video data, including the number of video samples per subject and scenario. While the dataset is limited to four subjects due to the inherent challenges and costs associated with recruiting and testing this vulnerable population, commonly leading to small sample sizes in highly specialized prosthetics research \cite{beaudette2018appropriateness, barnett2012small, hafner2016issues}, it's noteworthy that each subject presents highly diverse gait patterns and poses across various walking trials. This variability arises from the diverse prosthesis configurations, including differences in knee and ankle angles, pylon lengths, and knee mechanisms. These factors influence the subject's walking dynamics, ensuring diversity in video data samples. As a result, the dataset captures a broad spectrum of gait variations, which is crucial for training and evaluating vision models in real-world scenarios. The promising results achieved even with this small set of initial samples, on the other hand, underscores the potential of our methodology for future scaled-up studies,
which will certainly aim to incorporate broader demographic and prosthetic diversity.

\begin{table}[h!]
%	\vskip -0.1in
	\centering
	\begin{tabular}{c|c|c|c}
		\toprule
		Subject ID & Scenario & Trials & Samples \\
		\midrule
		\multirow{2}{*}{\textbf{P1}} & Inside Bars & 20 & 67 \\
		& Outside Bars & 32 & 95 \\
		\midrule
		\multirow{2}{*}{\textbf{P2}} & Inside Bars & 7 & 21 \\
		& Outside Bars & 34 & 100 \\
		\midrule
		\multirow{2}{*}{\textbf{P3}} & Inside Bars & 1 & 2 \\
		& Outside Bars & 9 & 25 \\
		\midrule
		\multirow{2}{*}{\textbf{P4}} & Inside Bars & 22 & 54 \\
		& Outside Bars & 19 & 48 \\
		\bottomrule
	\end{tabular}
	\caption{Basic Distribution of ProGait Dataset}
	\label{tab:dataset_statistics}
	\vskip -0.1in
\end{table}

\subsection{Annotations}
\label{annotation}

To provide high-quality annotations for the three tasks of using the ProGait dataset, we employed SOTA vision models to provide initial annotations in an automated manner, followed by manual refinements to ensure accuracy.

\begin{figure}[ht]
	%	\vskip -0.1in
	\begin{center}
		\includegraphics[width=1\columnwidth]{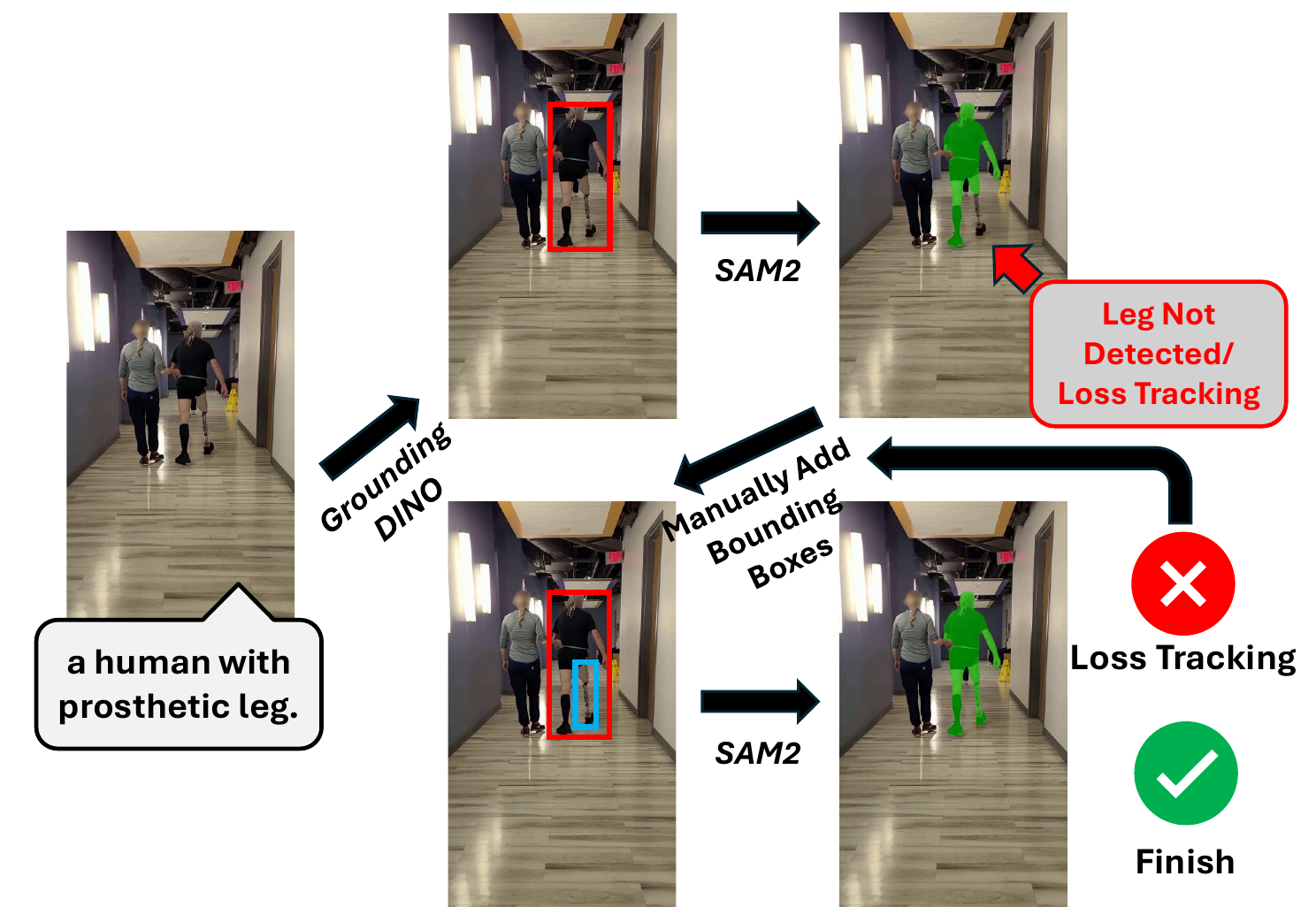}
		\caption{Pipeline of annotation }
		\label{fig:annotation}
	\end{center}
	\vskip -0.2in
\end{figure}

\noindent \textbf{Video Object Segmentation(VOS)}: There are three major challenges when utilizing the pre-trained vision models for tracking the subject and generating segmentation mask:

\begin{enumerate}[leftmargin=0.35in]
	\item Precisely detecting the prosthesis as part of the human body.
	\item Correctly distinguishing the subject from the healthcare staff.
	\item Ensuring consistent tracking despite occlusions.
\end{enumerate}

The second and third challenges are particularly pronounced in scenarios outside the parallel bars, where multiple people, including the healthcare professional, are present alongside the subject. 

As illustrated in Figure \ref{fig:annotation}, we address these challenges by developing a Human-in-the-Loop annotation pipeline with the help of Grounded SAM \cite{ren2024grounded}. The process begins with extracting the initial frame of the video and using the GroundingDINO model \cite{liu2024grounding} with the prompt ``a human with a prosthetic leg'' to annotate the subject's bounding box. Manual refinement is then performed to add additional bounding boxes, allowing the prosthesis to be tracked as a separate object. These bounding boxes are used as input for the Segment Anything Model 2 (SAM2) \cite{ravi2024sam}, which generates segmentation masks and tracks objects across frames. To ensure accuracy, we visually inspect for any loss of tracking. If the tracking fails, supplementary bounding boxes are added at the problematic frames, and segmentation masks are regenerated to maintain consistency.

Using the segmentation masks provided by SAM2, we also derive the final bounding boxes for each frame, ensuring consistency and accuracy throughout the sequence. This semi-automated approach reduces the need for frame-by-frame manual annotation while maintaining high annotation quality. The resulting annotations include precise bounding boxes and segmentation masks that facilitate robust object tracking and segmentation for prosthesis-specific tasks.

\begin{figure}
%	\vskip -0.1in
	\begin{center}
		\includegraphics[width=0.7\columnwidth]{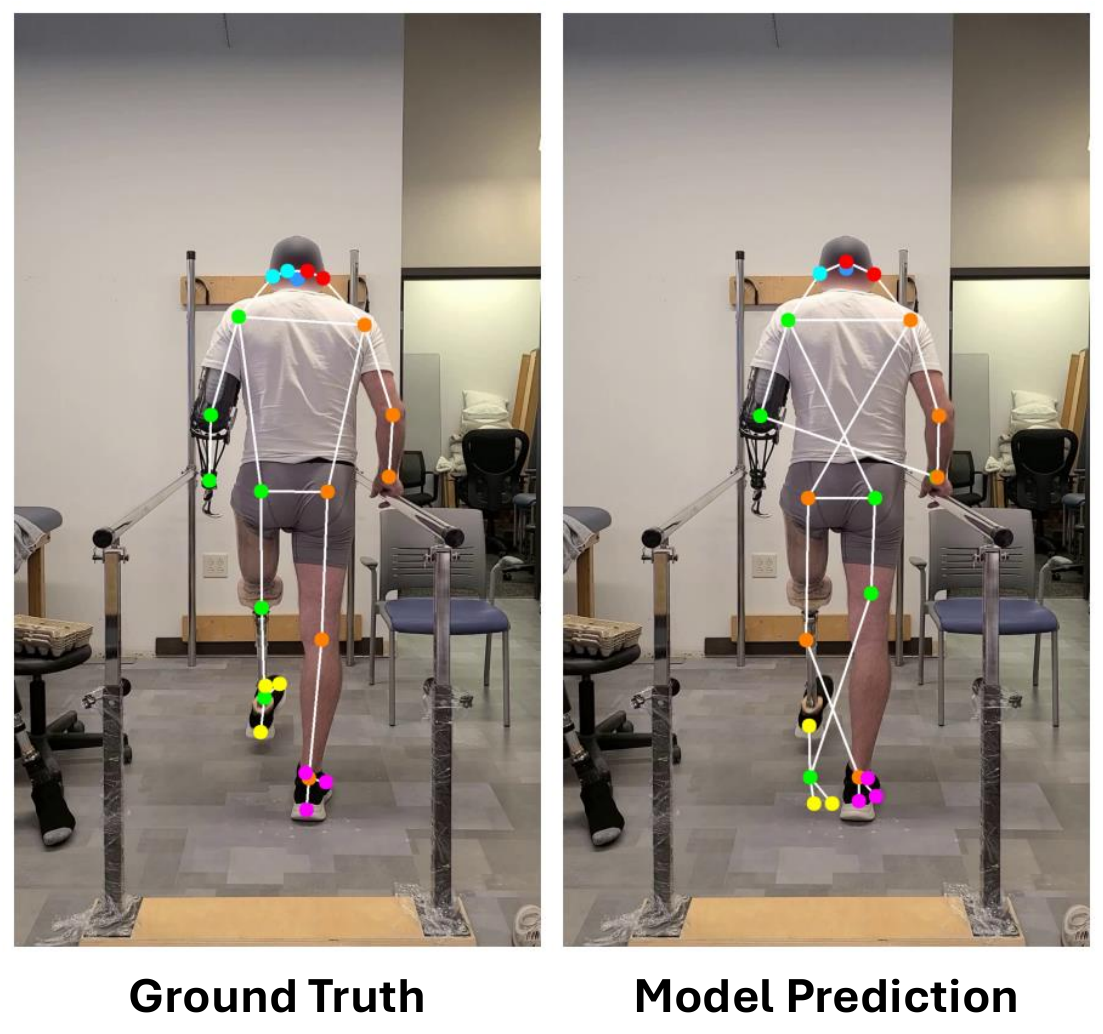}
%		\vskip -0.1in
		\caption{Zero-shot pose estimation with RTMW}
		\label{fig:pose_estimation}
	\end{center}
	\vskip -0.2in
\end{figure}

\noindent\textbf{2D Human Pose Estimation (HPE).} In this dataset, we focus on 23 pose keypoints, 17 for body and 6 for foot, defined in the COCO-WholeBody dataset \cite{jin2020whole}. After generating the segmentation mask, we apply it to the original videos and estimate the poses by applying the pre-trained RTMW model \cite{jiang2024rtmw} within the MMPose framework \cite{mmpose2020} on the masked video. However, the first challenge mentioned above remains unresolved. As illustrated in Figure \ref{fig:pose_estimation}, the pre-trained model frequently fails to accurately detect keypoints on the prosthetic knees and feet. Unlike the segmentation task, where the prosthesis can be tracked as an independent object, incorrect skeleton keypoints require labor-intensive, frame-by-frame manual correction.

To tackle this issue, we first randomly extract a small set of frames ($\sim$100) from the videos and manually correct the keypoints. We then fine-tune the RTMW model on these annotated frames and use the fine-tuned model to infer keypoints for a larger set of frames ($\sim$1,000). Next, we visually inspect the annotations from the fine-tuned model and select high-quality samples for a second round of fine-tuning. Finally, we apply the fine-tuned RTMW model to automatically annotate pose keypoints across all video sequences, manually correcting any remaining errors. After these two stages, only $<$25\% of the videos require further manual correction, significantly reducing the amount of workload for dataset preparation.

\begin{figure}[ht]
	\vskip -0.1in
	\begin{center}
		\includegraphics[width=0.98\columnwidth]{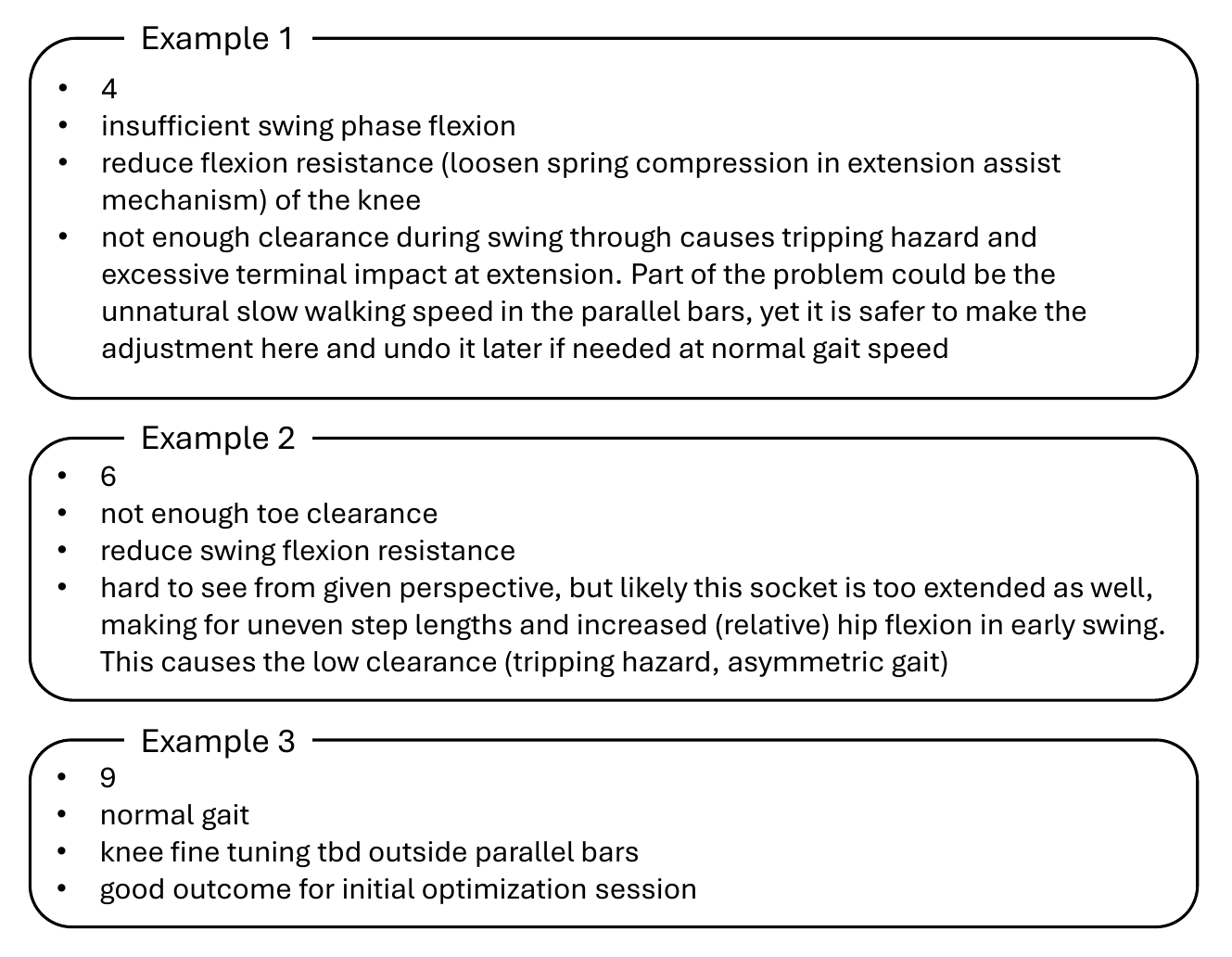}
		\vskip -0.1in
		\caption{Gait annotations}
		\label{fig:gait_annotation}
	\end{center}
	\vskip -0.2in
\end{figure}

\noindent\textbf{Gait Analysis (GA).} To facilitate gait analysis, we engaged researchers in rehabilitation sciences and human engineering to provide detailed textual descriptions for each video sample. As shown in Figure \ref{fig:gait_annotation}, these descriptions consist of four key components: (1) the general gait category, (2) the specific gait deviation, (3) recommendations on how to adjust the prosthesis to correct the gait, and (4) the reasons of these recommendations. 

While our benchmark task for GA primarily focuses on the classification problem over the general gait category, we provide full access to the other three components in the dataset, enabling advanced applications such as prosthesis optimization and clinical decision-making. In total, the annotations include 9 different gait categories, with each category comprising several fine-grained gait deviations. These deviations are classified according to their relevance to critical prosthetic alignment factors, such as knee and ankle positioning. Details of these categories are listed in Appendix \ref{sec:categories}, and the class distribution is shown in Table \ref{tab:class_distribution}.

\begin{table}[h!]
%	\vskip -0.1in
	\centering
	\hspace{-0.15in}
	\begin{tabular}{c|c|c}
		\toprule
		Code & Category & \# Samples \\
		\midrule
		1 & Rotational Deviations & 37 \\
		\midrule
		2 & Step and Base of Support Deviations & 41 \\
		\midrule
		3 & Step Length and Timing Issues & 13 \\
		\midrule
		4 & Knee Instability and Malalignment & 13 \\
		\midrule
		5 & Prosthetic Length Issues & 11 \\
		\midrule
		6 & Roll-Over and Clearance Issues & 6 \\
		\midrule
		7 & Socket Fit and Stability & 4 \\
		\midrule
		8 & Foot and Ankle Deviations & 8 \\
		\midrule
		9 & Normal Gait & 11 \\
		\bottomrule
	\end{tabular}
	\caption{Distribution of general gait categories. The hierarchical relationship between general gait categories and gait deviations can be found in the supplementary materials.}
	\label{tab:class_distribution}
	\vskip -0.1in
\end{table}

\subsection{De-identification}
To protect the privacy of individuals in the videos, we implement a comprehensive de-identification process for all video clips. Using a similar approach to segmentation mask annotation, we employ GroundingDINO and SAM2 models to detect sensitive elements, such as human faces and identifiable signage, in the first frames of video clips. We then manually add supplementary bounding boxes to cover these elements appearing later in the video. These identified areas are then subjected to Gaussian blurring to effectively obscure sensitive information. This semi-automated process ensures consistent and reliable de-identification while preserving the dataset's integrity for research purposes. Additionally, we conduct manual inspections to verify that all sensitive details are properly anonymized and that no identifiable features remain.

\section{Benchmarks and Baseline Models}
To facilitate the effective use of the ProGait dataset, we establish several benchmark tasks and provide fine-tuned baseline models as reference implementations. These benchmarks serve as an initial guide for researchers and practitioners, helping them evaluate the model performance across key tasks relevant to prosthetic gait analysis.

\subsection{Benchmark Tasks and Metrics}
%We use the three target tasks of the ProGait dataset, namely \emph{1)} Video Object Segmentation (VOS), \emph{2)} 2D Human Pose Estimation (HPE) and \emph{3)} Gait Classification, as the benchmark tasks. 
%Each task is designed to address a critical aspect of understanding and analyzing the movement patterns of individuals with transfemoral prosthetic legs.

\noindent\textbf{Video Object Segmentation.} For the segmentation task, we primarily use the mean Intersection over Union (mIoU) as the evaluation metric. The mIoU measures the overlap between the predicted mask and the ground-truth mask across video frames through a pixel-wise calculation, such that
\begin{equation}
	\text{mIoU}=\frac{1}{N}\sum_{i=1}^{N}\frac{\left | P_i\cap G_i\right |}{\left | P_i\cup G_i\right |}
\end{equation}
where $N$ is the total number of frames, $P_i$ is the predicted mask for frame $i$ and $G_i$ is the ground-truth mask for frame $i$. Intersection ($\cap$) measures the number of pixels that are correctly predicted as belonging to the class, and Union ($\cup$) measures the total number of pixels that are either in the predicted mask or the ground truth. This metric, hence, evaluates if the segmentation models can accurately delineate both the natural body parts and prosthetic components, which is crucial for downstream applications such as pose estimation and biomechanical analysis.

\noindent\textbf{2D Human Pose Estimation.} We evaluate the accuracy of pose estimation following the COCO-WholeBody standard \cite{jin2020whole}, and use the Average Precision (AP) across different Object Keypoint Similarity (OKS) thresholds ranging from 0.5 to 0.95, namely the AP@[.5,.95] as the primary metric. This metric can be presented as:
\begin{equation}
	\text{AP}@[0.5, 0.95] = \frac{1}{K} \sum_{t \in \{0.5, 0.55, ..., 0.95\}} AP_t
\end{equation}
where $K$ is the number of OKS thresholds (in this case, $K=10$ because the range is from 0.5 to 0.95 with a step of 0.05), and $AP_t$ is the Average Precision calculated by computing the area under precision-recall curves at a specific OKS threshold $t$. OKS measures the normalized distance between the predicted and ground-truth keypoints, defined as:
\begin{equation}
	\text{OKS} = \frac{\sum_{i} \exp\left(-\frac{d_i^2}{2s^2k_i^2}\right) \delta(v_i > 0)}{\sum_{i} \delta(v_i > 0)}
\end{equation}
where $d_i$ is the Euclidean distance between the $i$-th corresponding ground truth and the detected keypoint, $v_i$ is the visibility flag of the ground truth, $s$ is the object scale, $k_i$ is a per-keypont constant that controls falloff.

\noindent\textbf{Gait Classification.} This task categorizes video clips into predefined gait classes, by capturing the variations in movement patterns induced by different prosthetic configurations, walking conditions, or rehabilitation progress. We provide an end-to-end pipeline for training and evalua b ting different classification models. Performance is primarily measured using Top-1 accuracy, which indicates the proportion of correctly classified gait patterns. Balanced accuracy, defined as the average of sensitivity (recall) for each class, is also used as a reference to account for our unbalanced class distribution. 

Given that gait patterns can be subtle and complex, multiple gait deviations can, and often do, co-occur in clinical reality. However, in a clinical setting, physical therapists and prosthetists typically address one primary alignment issue at a time to avoid the problem of ``analysis paralysis'' \cite{jonkergouw2016effect, hashimoto2023proper}. This iterative process allows for precise adjustment and clear assessment of the impact of each modification, preventing the introduction of new problems while attempting to fix multiple simultaneously. Therefore, in our benchmark studies, we only consider the primary gait deviation for our classification task. 

\begin{figure}[ht]
	\vskip -0.1in
	\begin{center}
		\includegraphics[width=1.1\columnwidth]{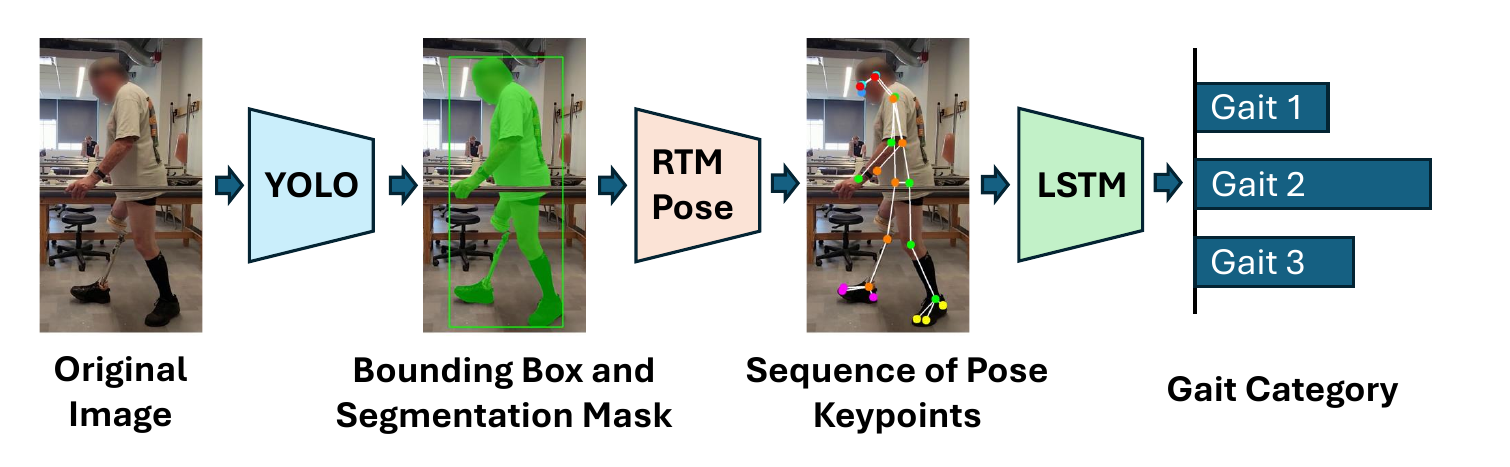}
		\vskip -0.1in
		\caption{The pipeline of baseline models}
		\label{fig:baseline_models}
	\end{center}
	\vskip -0.3in
\end{figure}

\subsection{Baseline Models}
\label{subsec:baselines}
To benchmark our dataset in the aforementioned tasks, we fine-tuned several vision models listed below as baselines. These models operate in a pipeline, as shown in Figure \ref{fig:baseline_models}, where the output of one model serves as the next's input. 
\begin{itemize} %[leftmargin=0.25in]
	\item For the Video Object Segmentation task, we use \textbf{YOLO11} \cite{yolo11_ultralytics}, the latest iteration in the Ultralytics YOLO series, capable of tracking multiple objects across frames. 
	\item For the 2D Human Pose Estimation task, we use \textbf{RTMPose} \cite{jiang2023rtmpose} from the MMPose framework. RTMPose takes the detected bounding boxes from the tracking model (YOLO11) as input and predicts pose keypoint positions.
	\item For the Gait Classification task, we use \textbf{A custom LSTM classifier}, which processes the sequence of poses over time and classifies the gait pattern for each video sample.
\end{itemize}

When fine-tuning the pre-trained models using the ProGait dataset, the dataset is divided into $\sim$70\% for training, $\sim$20\% for validation and $\sim$10\% for testing. To prevent data leakage, subjects appearing in the test set do not appear in the training or validation sets. 
%Instead of using every video frame for training, we sample one frame per second from each video to build the training and validation set. This speeds up the training, reduces redundancy, and focuses training on representative frames. 

Since the majority of our pose annotations, including those in the test set, are derived from the fine-tuned RTMW model, we refrain from using it as the baseline for the 2D Human Pose Estimation task. Instead, we use an RTMPose variant and fine-tune it exclusively on the training and validation sets. During the training of RTMPose and LSTM classifier, we use ground truth data as input, rather than the outputs from preceding models. Note that, RTMPose generates 133-point whole-body poses; however, we evaluate only the 23 keypoints corresponding to the body and feet, aligning with our annotations. Since gait patterns primarily involve the lower body, we exclude keypoints related to face and arms, and only 12 keypoints are used for the LSTM classifier.

\section{Experiments}
In this section, we present benchmark results by comparing our baseline models described in Section \ref{subsec:baselines} with other off-the-shelf models, for the aforementioned three tasks on the ProGait dataset.

\subsection{Experimental Setup}
To ensure a fair comparison, all evaluations are done only on the test set, and all the subjects in the test set are not present in either the training set or the validation set. This guarantees that the test samples are entirely unseen by the baseline models.Unlike the process of fine-tuning the baseline models described in Section \ref{subsec:baselines}, the experiment results are computed across all frames of the input videos rather than sparsely sampled frames. This approach provides a more comprehensive assessment of the model's consistency across frames.

\subsection{Video Object Segmentation (VOS)}
For the VOS task, we compare our fine-tuned YOLO11 model with the original checkpoint and the Grounded SAM2 with different text prompts. 

\begin{table}[ht]
%	\vskip -0.1in
	\centering
	\hspace{-0.16in}
	\begin{tabular}{l|c|c|c}
		\toprule
		Method & mIoU & \makecell{mIoU \\ (inside)} & \makecell{mIoU \\ (outside)} \\
		\midrule
		YOLO11 & 0.784 & 0.831 & 0.774 \\
		\midrule
		Grounded SAM2 & & & \\
		``a person.'' &0.358 &0.643 &0.559 \\
		``an amputee.'' & 0.905 &0.90 &0.907 \\
		\makecell[l]{``a human with\\prosthetic leg.''} &0.964 &0.921 &0.929 \\
		\midrule
		YOLO11-ProGait & 0.847 & 0.815 & 0.866 \\
		\bottomrule
	\end{tabular}
	\caption{Results for Video Object Segmentation}
	\label{tab:VOS_results}
	\vskip -0.1in
\end{table}

\begin{figure}[ht]
	\vskip -0.1in
	\begin{center}
		\includegraphics[width=0.55\columnwidth]{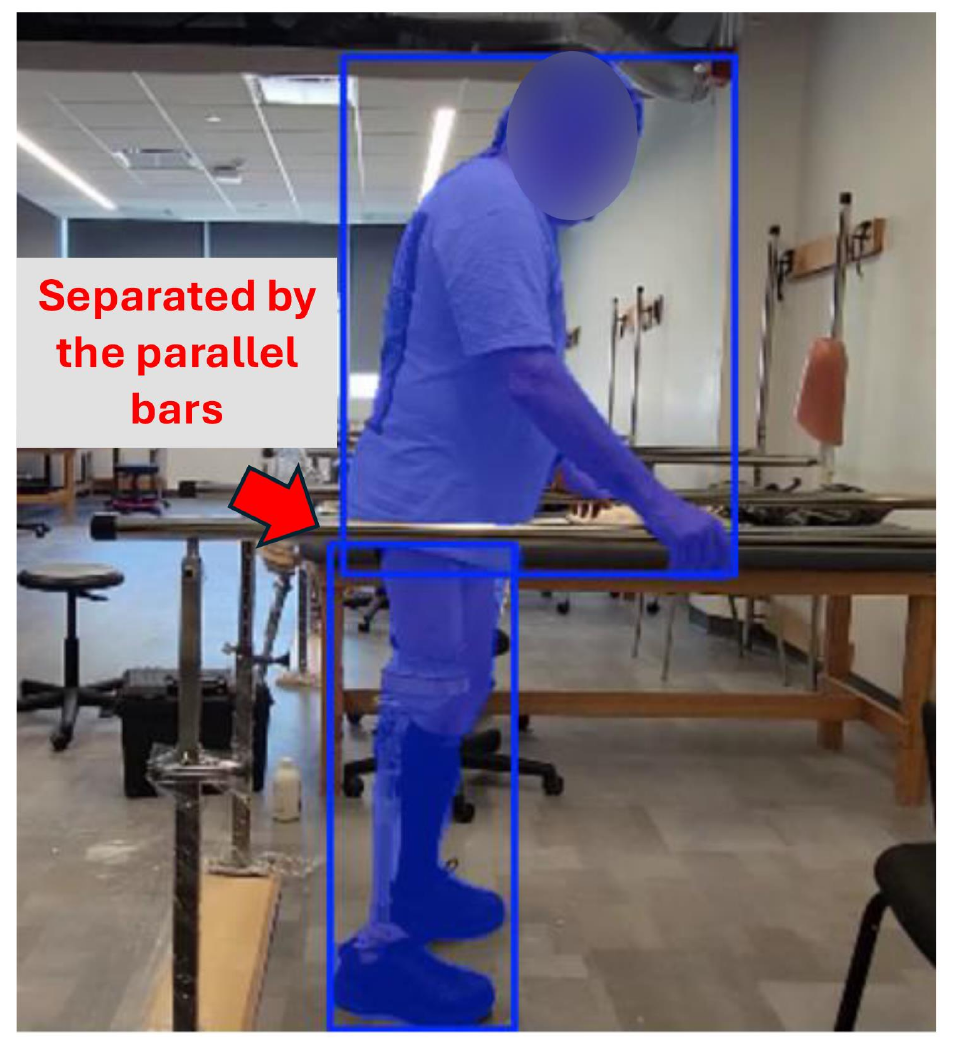}
		\caption{Fine-tuned YOLO may detect subject body as separate instances.}
		\label{fig:yolo_mask}
	\end{center}
	\vskip -0.2in
\end{figure}

Since the YOLO11 model is originally trained for multi-object detection, tracking and instance segmentation, it can produce multiple mask outputs. To evaluate its performance in tracking a single subject, we compute the mIoU using only the predicted mask with the largest intersection with the ground truth. On the other hand, the fine-tuned YOLO11 may occasionally detect the human body as separate instances, as illustrated in Figure \ref{fig:yolo_mask}. This occurs because YOLO11 can only be trained on closed shapes and cannot inherently handle discrete parts of a single subject. To address this problem, we train the YOLO11 model to learn discrete parts separately and merge all detected mask instances into a single unified mask for evaluation.

The results in Table \ref{tab:VOS_results} show that, Grounded SAM2 with an appropriate text prompt can achieve very good performance, but highly depends on the specific text prompt being used. Further, even with mIoU up to 96\%, it can occasionally loss tracking of the prosthetic leg, as shown in Figure \ref{fig:sota_limitation}. On the other hand, although the issue of detecting body parts separately causes a slight accuracy drop in scenario of inside parallel bars, YOLO11-ProGait generally outperforms its original checkpoint. Given that prosthetic legs are typically thin and occupy a small area in the frame, the performance gap between the pre-trained and fine-tuned models remains modest, which is reasonable.

\subsection{2D Human Pose Estimation (HPE)}
In the 2D HPE task, for comparison with our fine-tuned RTMPose model, we selected three pre-trained models for 2D whole-body pose estimation: HRNet \cite{Sun_2019_CVPR}, ViPNAS \cite{xu2021vipnas}, and ViTPose \cite{xu2022vitpose} pre-trained on COCO-WholeBody dataset\cite{jin2020whole}, apart from the original RTMPose checkpoint.

\begin{table}[ht]
%	\vskip -0.1in
	\centering
	\hspace{-0.16in}
	\begin{tabular}{l|c|c|c}
		\toprule
		Method & AP & \makecell{AP \\ (inside)} & \makecell{AP \\ (outside)} \\
		\midrule
		HRNet \cite{Sun_2019_CVPR} &0.750 &0.825 &0.733 \\
		ViPNAS \cite{xu2021vipnas} &0.761 &0.735 &0.767 \\
		RTMPose \cite{jiang2023rtmpose}& 0.855 &0.876 &0.850 \\
		\midrule
		ViTPose \cite{xu2022vitpose}& 0.830 & 0.845 & 0.822 \\
		\midrule
		RTMPose-ProGait & 0.947 &0.968 &0.942 \\
		\bottomrule
	\end{tabular}
	\caption{Results for 2D Human Pose Estimation}
	\label{tab:HPE_results}
	\vskip -0.1in
\end{table}

We first evaluated the $AP@[0.5, 0.95]$ metric for all the 23 keypoints (17 keypoints for ViTPose\footnote{Due to its adherence to the standard COCO 17-keypoints format, ViTPose cannot detect the 6 feet points.}) , and the results are shown in Table \ref{tab:HPE_results}. Our baseline model RTMPose-ProGait outperforms all the other pre-trained models, demonstrating the effectiveness of using the ProGait dataset to improve the prosthesis detection.

Additionally, we evaluated the AP for lower body keypoints. More specifically, we calculate the score for 10 out of the 23 keypoints, covering knees, ankles, and feet. ViTPose was excluded due to its lack of 6 feet points. The results in Table \ref{tab:HPE_results_leg} shows that our RTMPose-ProGait significantly performs  better than other methods on this metric. On the one hand, these results also show that the existing models struggle in detecting prostheses as part of the human body, and the ProGait dataset can well address this problem.

\begin{table}[ht]
%	\vskip -0.1in
	\centering
	\hspace{-0.16in}
	\begin{tabular}{l|c|c|c}
		\toprule
		Method & AP-leg & \makecell{AP-leg \\ (inside)} & \makecell{AP-leg \\ (outside)} \\
		\midrule
		HRNet & 0.625 &0.691 &0.539 \\
		ViPNAS & 0.631 &0.527 &0.655 \\
		RTMPose & 0.804 &0.761 &0.814. \\
		\midrule
		RTMPose-ProGait &0.918 &0.918  &0.918  \\
		\bottomrule
	\end{tabular}
	\caption{Results for 2D Human Pose Estimation with AP calculated on knee, ankle, and foot points only}
	\label{tab:HPE_results_leg}
%	\vskip -0.1in
\end{table}

\subsection{Gait Classification}

\begin{figure}[ht]
%	\vskip -0.1in
	\begin{center}
		\includegraphics[width=0.8\columnwidth]{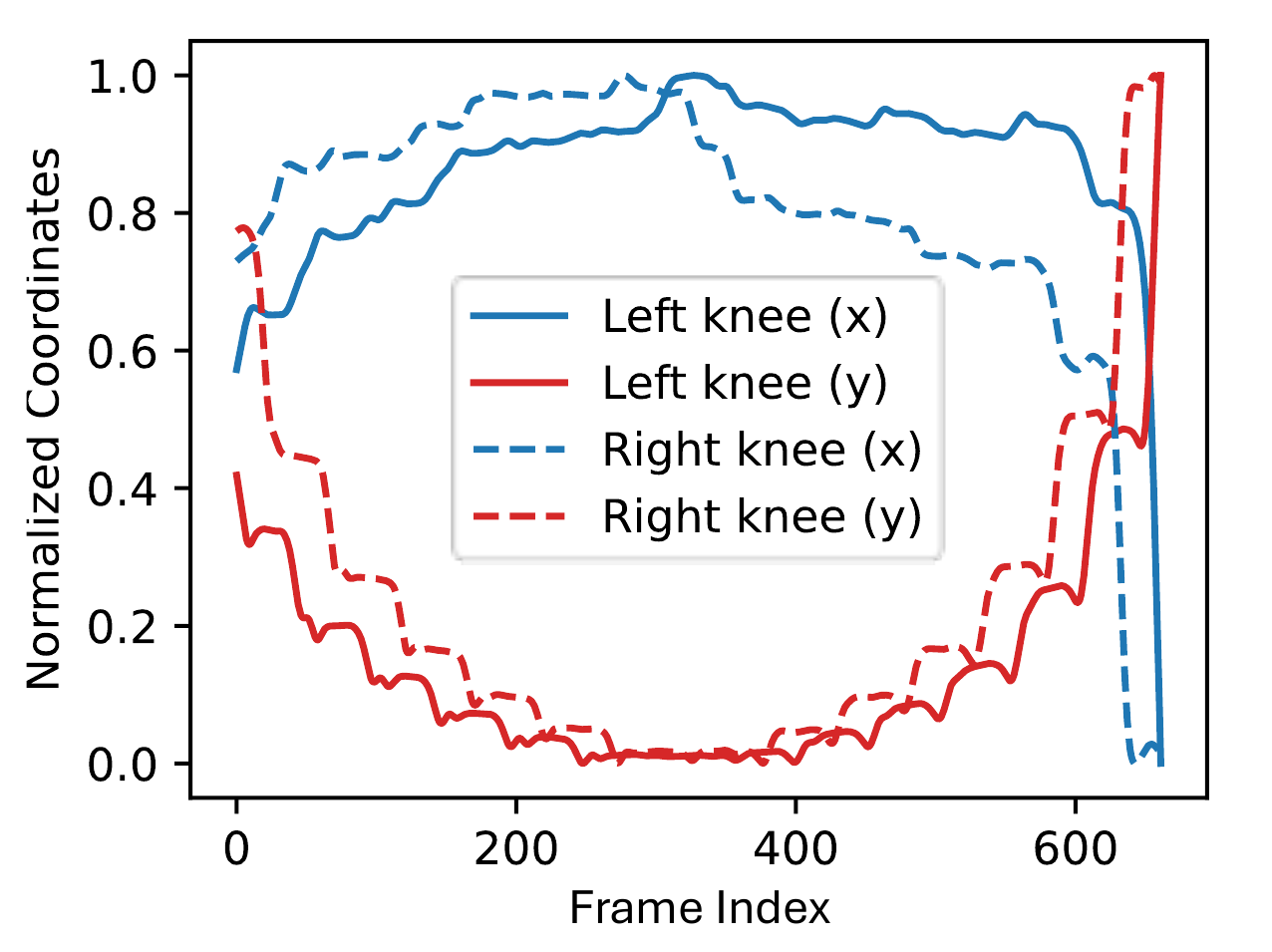}
		\caption{The sequence of pose keypoint coordinates}
		\label{fig:pose_sequences}
	\end{center}
	\vskip -0.1in
\end{figure}

To bridge the gap between pose sequences and gait analysis, we designed and trained a custom LSTM network with 128 hidden dimensions to process pose sequences as input. As an example shown in Figure \ref{fig:pose_sequences}, the $(x, y)$ coordinates of the pose keypoints show periodic motion over time. We utilize such time-series data and use it to classify the poses into 9 general gait categories. Since no off-the-shelf model exists for this specific task, we conducted a 5-fold cross-validation and evaluated the model across five different setups, as shown in Table \ref{tab:gait_classification}.

\begin{table}[ht]
%	\vskip -0.1in
	\centering
	\hspace{-0.16in}
	\begin{tabular}{l|c|c}
		\toprule
		 & Top-1 Acc. & Balanced Acc. \\
		\midrule
		Frontal view & 0.510 & 0.545  \\
		Sagittal view & 0.826 & 0.790  \\
		\midrule
		Inside parallel bars & 0.364 & 0.437  \\
		Outside parallel bars & 0.486 & 0.320  \\
		\midrule
		Whole dataset & 0.372 & 0.403  \\
		\bottomrule
	\end{tabular}
	\caption{Results for Gait Classification, with all 23 keypoints taken as input}
	\label{tab:gait_classification}
	\vskip -0.1in
\end{table}

Notably, the LSTM network performs exceptionally well when provided with pose sequences from the sagittal view alone. However, when both frontal and sagittal views are fed into the model, accuracy drops significantly. This suggests that treating pose sequences from both views simultaneously may introduce confusion. Additionally, it indicates that the sagittal view may be a more suitable angle for observing gait patterns.

We further compared the gait classification performance using all 23 keypoints vs. using only the 12 keypoints corresponding to the lower body. As shown in Table \ref{tab:gait_classifcation_lowerbody}, excluding upper-body keypoints has minimal impact on classification accuracy. While upper-body movements may introduce variations in gait patterns, gait is fundamentally driven by lower-body dynamics. This suggests that lower-body keypoints alone are sufficient for gait-related downstream tasks.

\begin{table}[ht]
	%	\vskip -0.1in
	\centering
	\hspace{-0.16in}
	\begin{tabular}{l|c|c}
		\toprule
		& Top-1 Acc. & Balanced Acc. \\
		\midrule
		Frontal view & 0.474 & 0.521  \\
		Sagittal view & 0.773 & 0.812  \\
		\midrule
		Inside parallel bars & 0.364 & 0.457  \\
		Outside parallel bars  & 0.566 & 0.423  \\
		\midrule
		Whole dataset & 0.384 & 0.413  \\
		\bottomrule
	\end{tabular}
	\caption{Results for Gait Classification, with lower-body keypoints taken as input only}
	\label{tab:gait_classifcation_lowerbody}
	\vskip -0.1in
\end{table}

For additional demonstration of ProGait's generality, we performed more experiments over multiple gait recognition methods, including GaitGraph2\cite{teepe2022towards}, GaitBase\cite{fan2023opengait}, and GPGait\cite{fu2023gpgait}, by applying our dataset onto the pre-trained model and using our dataset to fine-tune the pre-trained models. For the pre-trained model, we used the generated gait embeddings for 1-nearest-neighbor classification. In fine-tuning, we appended a classification head of two FC layers to the pre-trained model, and only trained these two layers. We also conducted experiments on ScoNet \cite{zhou2024gait}, a silhouette-based clinical gait classification model for scoliosis detection. We trained the model from scratch using the same original settings, and adapted its architecture by changing the output dimension from 3 to 9 to align with our 9-class classification task.

Results in Table \ref{tab:gait_classification_others} show that these gait recognition/classification methods achieve reasonable performance on the gait classification task, and with simple fine-tuning/retraining, they can be well-adapted to our ProGait dataset with significantly higher accuracy. These results also showed that our simple LSTM classifier provides competitive performances.

\begin{table}[ht]
	%	\vskip -0.1in
	\centering
	\hspace{-0.16in}
	\begin{tabular}{l|c|c}
		\toprule
		& Pre-trained & \makecell{Fine-tuned/ \\ Re-trained } \\
		\midrule
		GaitGraph2 \cite{teepe2022towards} & 0.200 & 0.440 \\
		GPGait \cite{fu2023gpgait} & 0.388 & 0.457 \\
		GaitBase \cite{fan2023opengait} & 0.313 & 0.340 \\
		\midrule
		ScoNet & N/A & 0.333 \\
		\midrule
		LSTM-ProGait & N/A & 0.372 \\
		\bottomrule
	\end{tabular}
	\caption{Results for Gait Classification on the whole dataset, measured in Top-1 accuracies.}
	\label{tab:gait_classification_others}
\end{table}

\section{Discussion and Conclusion}
In this paper, we introduce ProGait, a versatile dataset designed to advance vision-based models for detecting lower-limb prostheses, facilitate comprehensive gait analysis to aid in prosthesis design and alignment, and support the development of assistive technologies that enhance mobility and overall quality of life for prosthesis users. By providing high-quality data tailored for multiple applications, ProGait aims to bridge the gap between computer vision research and real-world clinical needs, fostering innovations in prosthetic engineering and rehabilitation.

Beyond the tasks introduced in this paper, the textual descriptions of gait patterns and their corresponding reasoning unlock vast possibilities for the ProGait dataset. In future work, we plan to leverage this textual information alongside large language models (LLMs), to provide prosthesis users with quick and convenient assessments of stability, comfort, and mobility, while also reducing the workload of healthcare professionals. Ultimately, our goal is to enable the development of adaptive, interactive prosthetic legs powered by microcomputers, paving the way for intelligent, user-responsive prosthetic solutions.

{
	\small
	\bibliographystyle{ieeenat_fullname}
	\bibliography{main}
}

\clearpage
\appendix

\section{List of General Gait Category and Subcategories for Gait Deviation}
\label{sec:categories}
\begin{itemize}
	\item \textbf{Rotational Deviations:} These involve abnormal inward or outward twisting of the prosthetic limb or foot during walking, often leading to inefficient movement, joint stress, and potential discomfort or skin issues. They disrupt the natural alignment and progression of the leg through the gait cycle. \\ \textit{Subcategories}:
	\subitem Medial whip
	\subitem Lateral whip
	\subitem Toe-in asymmetry
	\subitem Internally rotated foot
	\subitem Toe-out asymmetry
	\item \textbf{Step and Base of Support Deviations:} This category covers issues with the width of the walking path or how the feet are placed, often indicating problems with stability, balance, or confidence. Such deviations can result in an unsteady, energy-intensive, or overly rigid gait. \\ \textit{Subcategories}:
	\subitem Too narrow step width
	\subitem Too narrow base of support
	\subitem Too wide base of support
	\subitem Abducted gait
	\subitem Leaning pylon
	\item \textbf{Step Length and Timing Issues:} These are inconsistencies in the length or synchronized timing of individual steps, typically signaling imbalances, pain, or problems with prosthetic control. They lead to an asymmetrical and less efficient walking pattern. \\ \textit{Subcategories}:
	\subitem Asymmetric step length
	\subitem Terminal impact
	\subitem Early foot flat
	\item \textbf{Knee Instability and Malalignment:} This category describes issues where the prosthetic knee is unstable or improperly aligned, significantly impacting safety and function, especially for above-knee amputees. It can cause a feeling of insecurity, increase fall risk, and lead to abnormal loading. \\ \textit{Subcategories}:
	\subitem Insufficient knee flexion
	\subitem Hyperextended knee
	\subitem Knee varus asymmetry
	\subitem Excessive valgus
	\subitem Externally rotated knee
	\item \textbf{Prosthetic Length Issues:} These deviations occur when the prosthetic limb is not the correct length, forcing the user to adopt compensatory movements throughout the body. Even minor discrepancies can drastically affect gait mechanics, increasing energy expenditure and potential for secondary problems. \\ \textit{Subcategories}:
	\subitem Prosthesis too short
	\subitem Prosthesis too long
	\subitem Hip drop
	\item \textbf{Roll-Over and Clearance Issues:} This category addresses problems with the smooth weight transfer over the prosthetic foot during stance or the ability of the prosthetic limb to clear the ground during swing. These issues greatly reduce walking fluidity and often trigger compensatory actions like hip hiking or circumduction. \\ \textit{Subcategories}:
	\subitem Incomplete roll-over
	\subitem Insufficient toe clearance
	\subitem Circumduction
	\item \textbf{Socket Fit and Stability:} These deviations are direct results of an ill-fitting or unstable prosthetic socket, which is crucial for comfort and function. Poor fit can cause pain, skin irritation, and instability, leading to compensatory movements and an inefficient gait. \\ \textit{Subcategories}:
	\subitem Socket too wide in ML
	\subitem Lateral instability
	\item \textbf{Foot and Ankle Deviations:} This category includes abnormalities in the prosthetic foot and ankle's position or movement, affecting shock absorption, propulsion, and balance. Such issues often stem from incorrect component selection, alignment, or the user's compensatory habits. \\ \textit{Subcategories}:
	\subitem Incongruity of knee and ankle axes
	\subitem Excessive plantar flexion
	\subitem Excessive dorsiflexion
	\subitem Pronated or everted foot
	\item \textbf{Normal Gait:} This describes an efficient, symmetrical, and coordinated walking pattern where muscles, joints, and the nervous system work in harmony for stability and propulsion. The goal for prosthetic users is often to achieve a gait as close to this natural, energy-minimizing pattern as possible.
\end{itemize}

\section{Additional Video Samples}
\label{sec:video_samples}
In this appendix, we provide additional video samples and visualized annotations in all scenarios and views. This provides a comprehensive view of the whole dataset, and shows the diversity of scenes, poses and gaits in ProGait dataset.

\onecolumn

\begin{figure}
%	\vskip -0.1in
	\begin{center}
		\includegraphics[width=1\columnwidth]{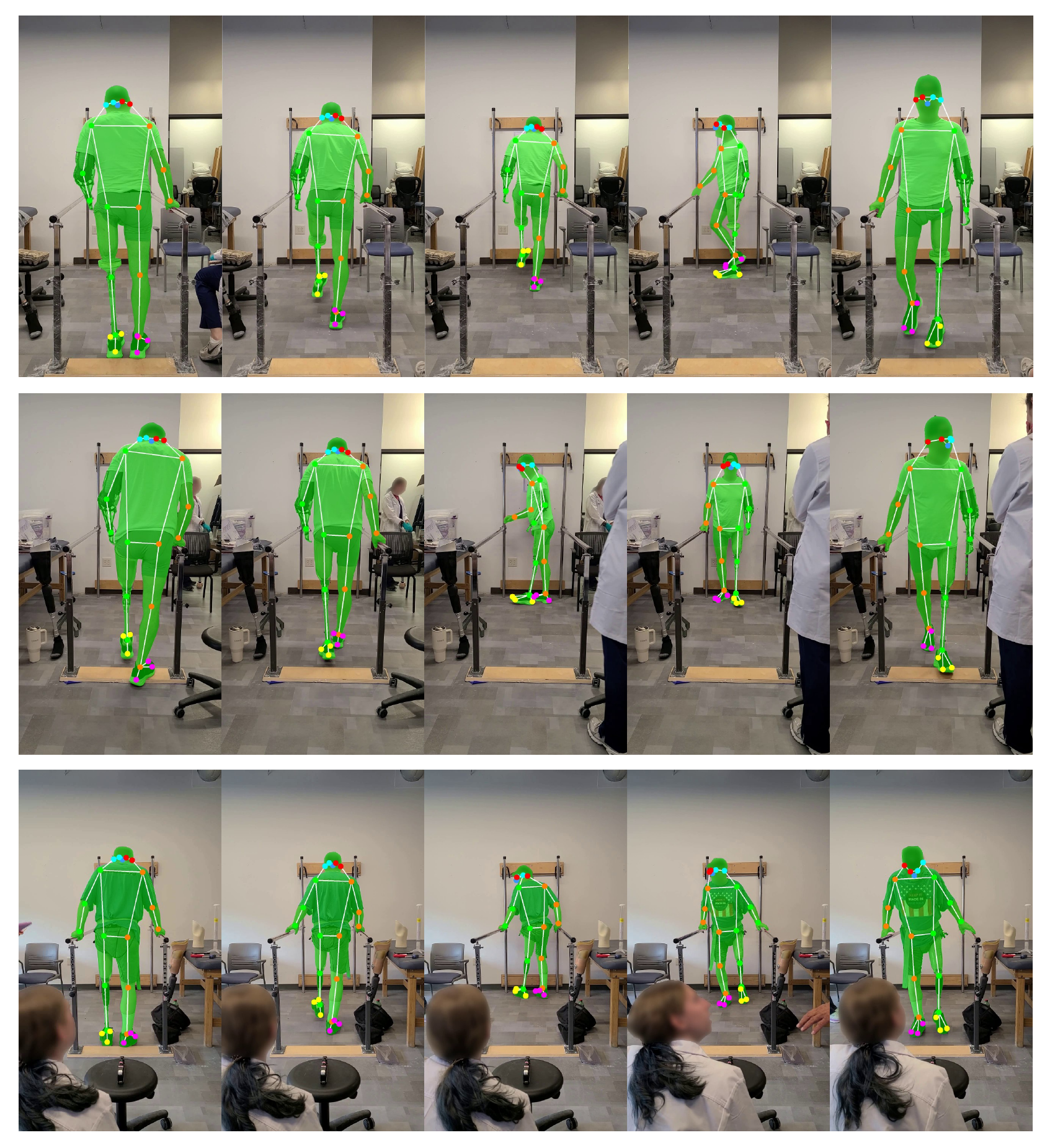}
		\caption{Videos and corresponding annotations captured at \textbf{frontal} view \textbf{inside} the parallel bar. Sequence 1 \& 2 shows Subject 1 walking with a simple mechanical knee vs. a hydraulic knee. Sequence 3 shows Subject 2 walking with another model of hydraulic knee.}
		\label{fig:sample1}
	\end{center}
	\vskip -0.1in
\end{figure}

\begin{figure}
%	\vskip -0.1in
	\begin{center}
		\includegraphics[width=1\columnwidth]{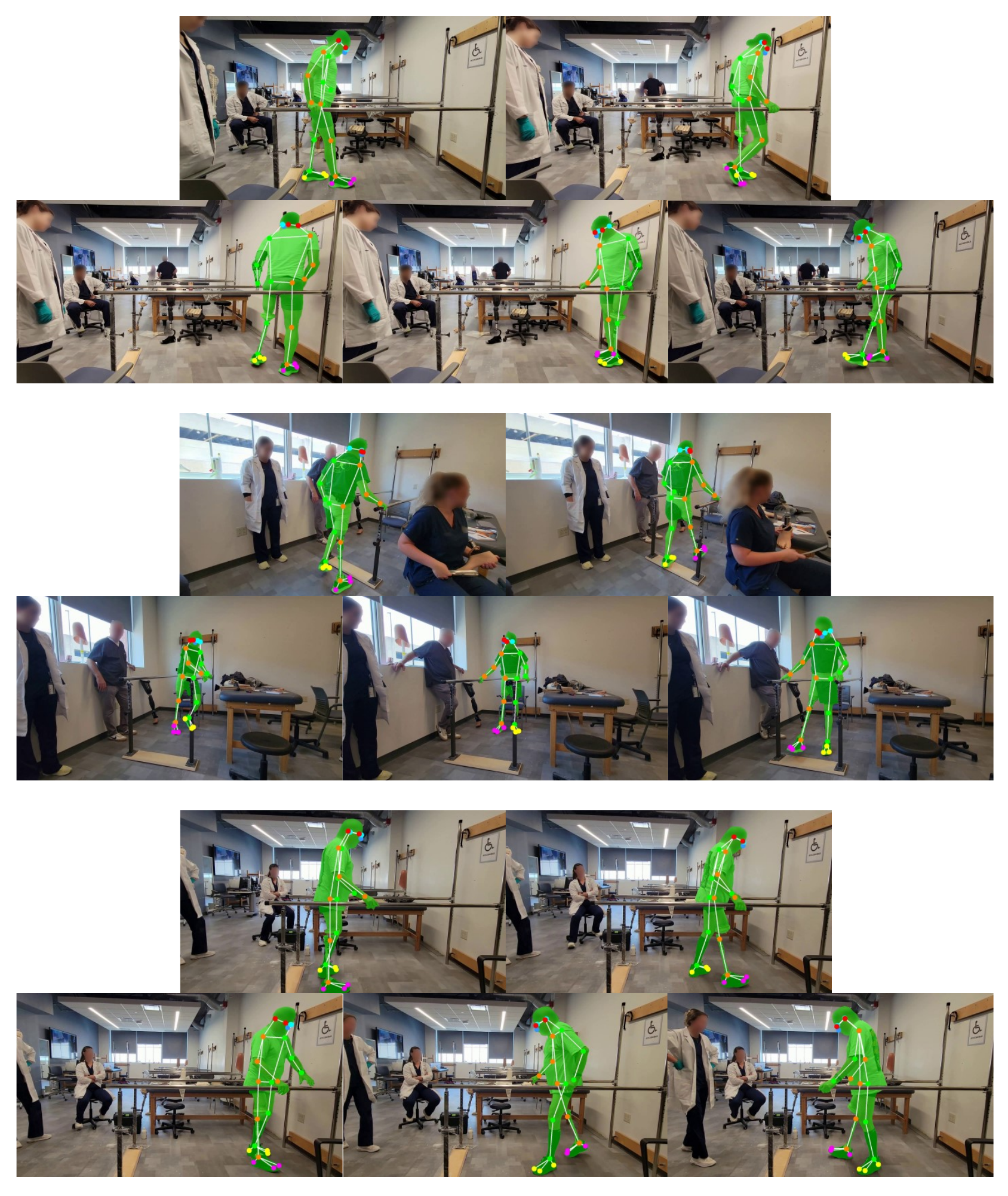}
		\caption{Videos and corresponding annotations captured at \textbf{sagittal} view \textbf{inside} the parallel bar. Subjects 1, 3, and 4 are shown walking with various prosthetic knee types: mechanical, hydraulic, and mechanical, respectively.}
		\label{fig:sample2}
	\end{center}
	\vskip -0.1in
\end{figure}

\begin{figure}
%	\vskip -0.1in
	\begin{center}
		\includegraphics[width=1\columnwidth]{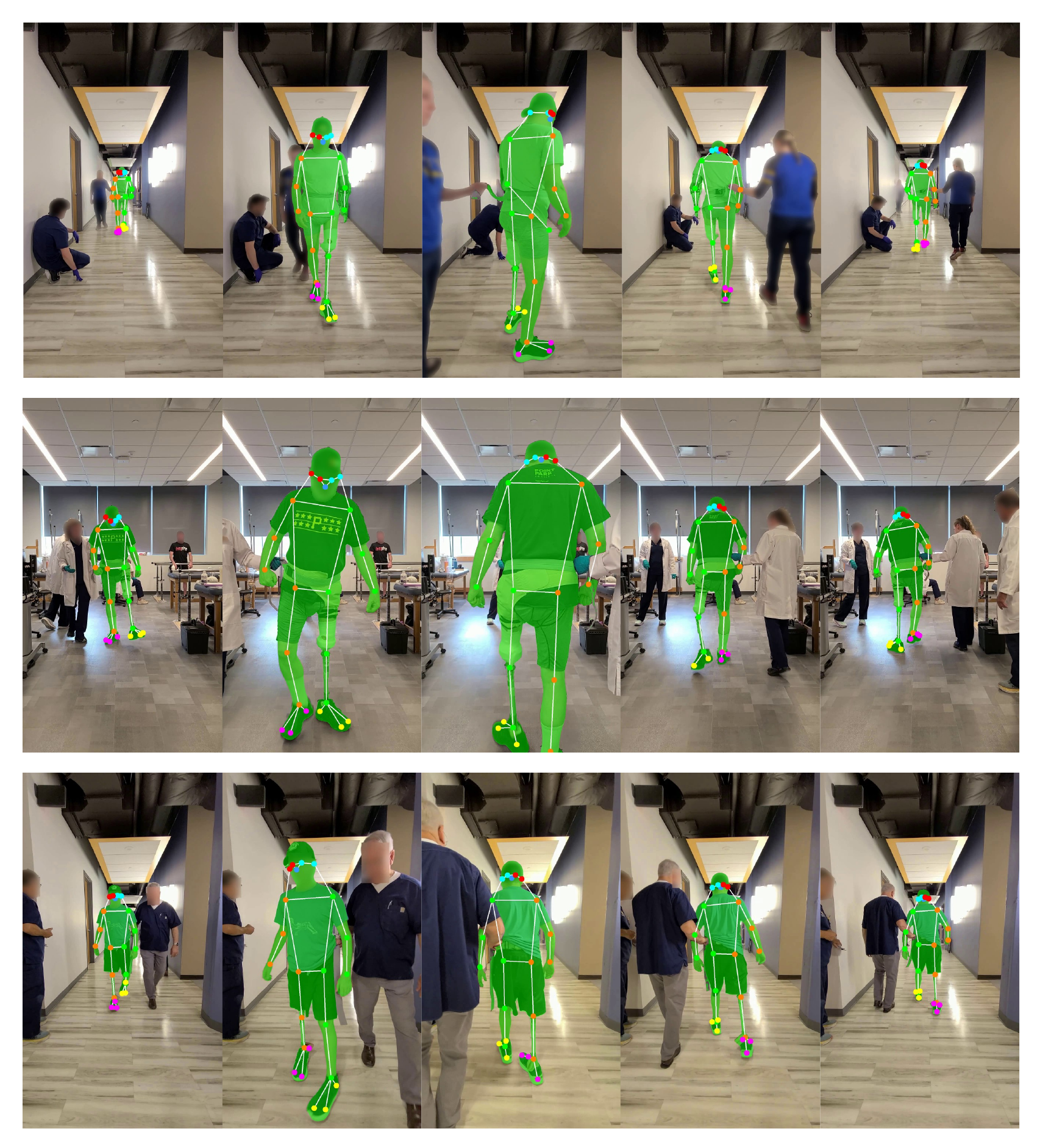}
		\caption{Videos and corresponding annotations captured at \textbf{frontal} view \textbf{outside} the parallel bar. Subject 1, 2, and 3 walking alongside the hallway with mechanical knees, accompanied by the healthcare staff.}
		\label{fig:sample3}
	\end{center}
	\vskip -0.1in
\end{figure}

\begin{figure}
%	\vskip -0.1in
	\begin{center}
		\includegraphics[width=1\columnwidth]{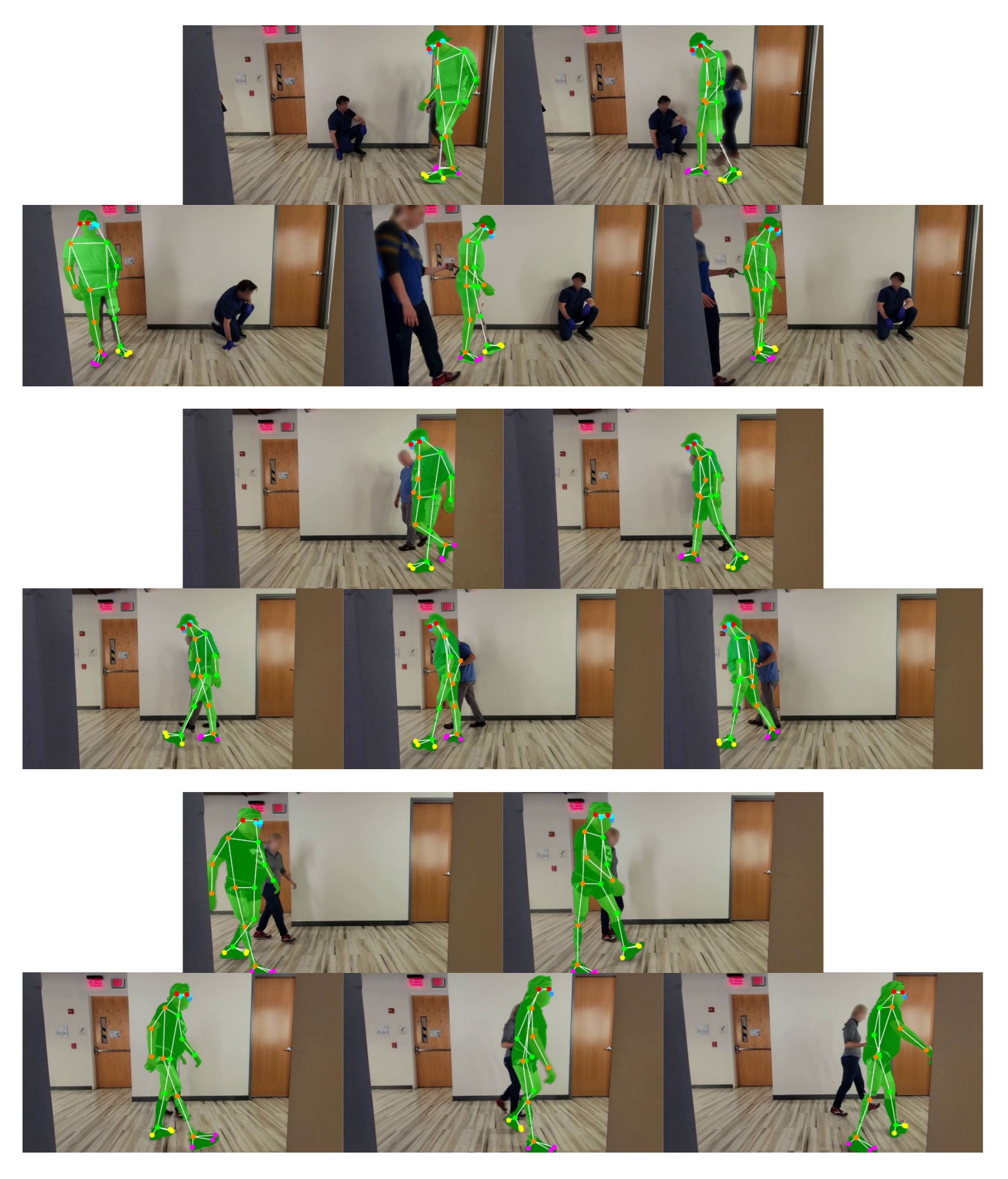}
		\caption{Videos and corresponding annotations captured at \textbf{sagittal} view \textbf{outside} the parallel bar. Subjects 1, 2, and 4 walking alongside the hallway with mechanical knees, accompanied by the healthcare staff.}
		\label{fig:sample4}
	\end{center}
	\vskip -0.1in
\end{figure}

\end{document}